\documentclass[letterpaper, 10 pt, conference]{ieeeconf}  

\IEEEoverridecommandlockouts                              

\usepackage{
    amsbsy, amsfonts, amsmath, amssymb, bm, graphicx,
    multicol, mwe, siunitx, subcaption, xcolor, pdfpages, cuted, titlesec, float, booktabs, hyperref, balance
}

\title{\LARGE \bf
Humanoid-Gym: Reinforcement Learning for Humanoid Robot with Zero-Shot Sim2Real Transfer
}

\author{Xinyang Gu$^{2*}$, Yen-Jen Wang$^{13*}$, Jianyu Chen$^{123}$
\thanks{$^*$Equal contribution. Listed alphabetically.}%
\thanks{$^{1}$Shanghai Qi Zhi Institute, Shanghai, China.}%
\thanks{$^{2}$RobotEra TECHNOLOGY CO., LTD., Beijing, China}%
\thanks{$^{3}$Tsinghua University, Beijing, China.}%
}

\begin{document}

\maketitle
\thispagestyle{empty}
\pagestyle{empty}

\begin{abstract}

Humanoid-Gym is an easy-to-use reinforcement learning (RL) framework based on Nvidia Isaac Gym, designed to train locomotion skills for humanoid robots, emphasizing zero-shot transfer from simulation to the real-world environment. Humanoid-Gym also integrates a sim-to-sim framework from Isaac Gym to Mujoco that allows users to verify the trained policies in different physical simulations to ensure the robustness and generalization of the policies. This framework is verified by RobotEra's XBot-S (1.2-meter tall humanoid robot) and XBot-L (1.65-meter tall humanoid robot) in a real-world environment with zero-shot sim-to-real transfer. The project website and source code can be found at: \href{https://sites.google.com/view/humanoid-gym/}{sites.google.com/view/humanoid-gym}.

\end{abstract}

\section{INTRODUCTION}

Modern environments are primarily designed for humans. Therefore, humanoid robots, with their human-like skeletal structure, are especially suited for tasks in human-centric environments, offering unique advantages over other types of robots. Recently, massively parallel deep reinforcement learning (RL) in simulation has become popular \cite{rudin2022learning,kumar2021rma,guo2023decentralized}. However, due to the complex structure of humanoid robots, the sim-to-real gap \cite{peng2018sim,zhao2020sim,tan2018sim,kadian2020sim2real} exists and is larger than that of quadrupedal robots. Therefore, we release Humanoid-Gym, an easy-to-use RL framework based on Nvidia Isaac Gym \cite{makoviychuk2021isaac}, designed to train locomotion skills for humanoid robots, emphasizing zero-shot transfer from simulation to the real-world environment. Humanoid-Gym features specialized rewards and domain randomization techniques for humanoid robots, simplifying the difficulty of sim-to-real transfer.
Furthermore, it also integrates a sim-to-sim framework from Isaac Gym \cite{makoviychuk2021isaac} to MuJoCo \cite{todorov2012mujoco} that allows users to verify the trained policies in different physical simulations to ensure the robustness and generalization of the policies, shown in Fig.~\ref{fig:envs}. Currently, Humanoid-Gym is verified by multiple humanoid robots with different sizes in a real-world environment with zero-shot sim-to-real transfer, including RobotEra's XBot-S (1.2-meter tall humanoid robot) and XBot-L (1.65-meter tall humanoid robot) \cite{robotera}. The contribution of Humanoid-Gym can be summarized as follows:

\begin{figure}[tp]
    \centering
    \includegraphics[width=1.0\linewidth]{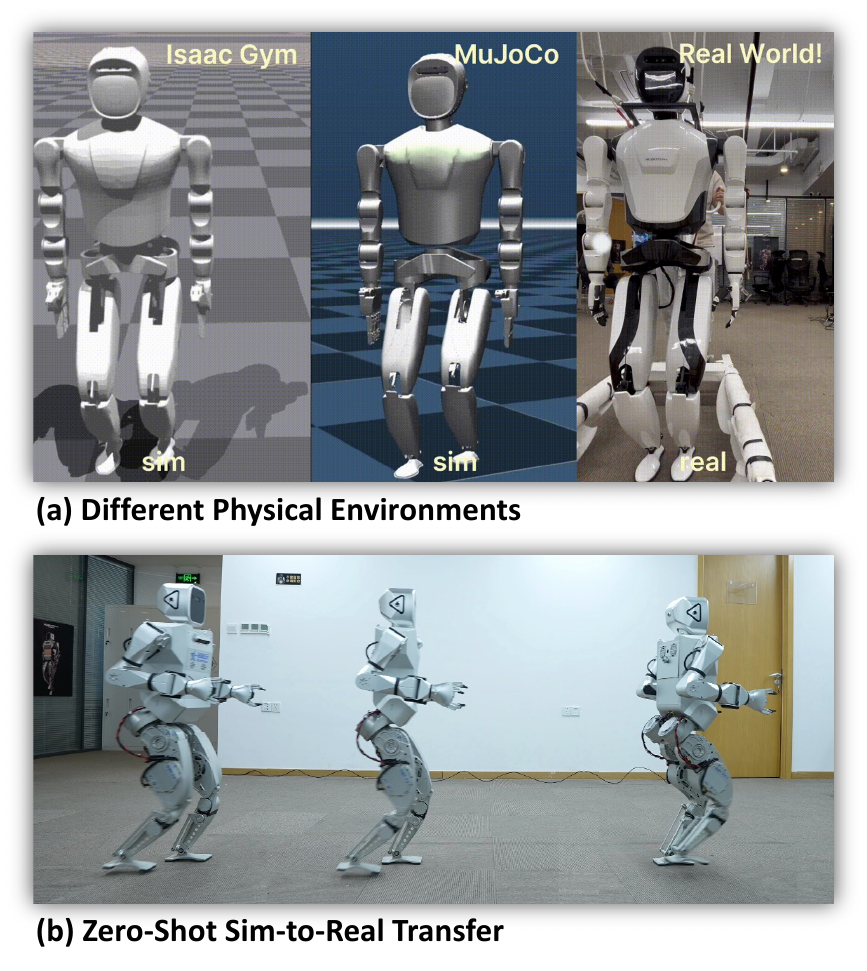}
    \caption{Humanoid-Gym enables users to train their policies within Nvidia Isaac Gym and validate them in MuJoCo. Additionally, we have successfully tested the complete pipeline with two humanoid robots. They were trained in Humanoid-Gym and transferred to real-world environments in a zero-shot manner.}
    \label{fig:envs}
\end{figure}

\begin{itemize}
    \item We launch an open-source reinforcement learning (RL) framework with meticulous system design.
    \item Our framework enables zero-shot transfer from simulation to the real world, which has been rigorously tested across humanoid robots of various sizes.
    \item Our open-source library features a sim-to-sim validation tool, enabling users to test their policies across diverse environmental dynamics rigorously.
\end{itemize}
\section{Related Works}
\subsection{Robot Learning on Locomotion Tasks}
Reinforcement learning (RL) has shown promise in enabling robots to achieve stable locomotion \cite{tan2018sim, hwangbo2019learning, lee2020learning}. Compared to prior RL efforts with quadrupedal robots \cite{rudin2022learning,margolis2022rapid} and bipedal robots like Cassie \cite{li2021reinforcement, kumar2022adapting}, our work with humanoid robots introduces a more challenging scenario for robot control. Recent studies \cite{radosavovic2023learning, radosavovic2024humanoid} have applied transformer architecture to improve the walking performance of humanoid robots on flat surfaces. Beyond lower-body control, some works \cite{he2024learning, cheng2024expressive} have also explored more complex upper-body skills for humanoid robot control. However, the sim-to-real transition for humanoid locomotion remains a significant challenge, with a notable lack of open-source resources in the robot learning community. To contribute to this area, we have developed Humanoid-Gym, an accessible framework with full codebase.
\section{Method}

\begin{figure}[tp]
    \centering
    \includegraphics[width=1\linewidth]{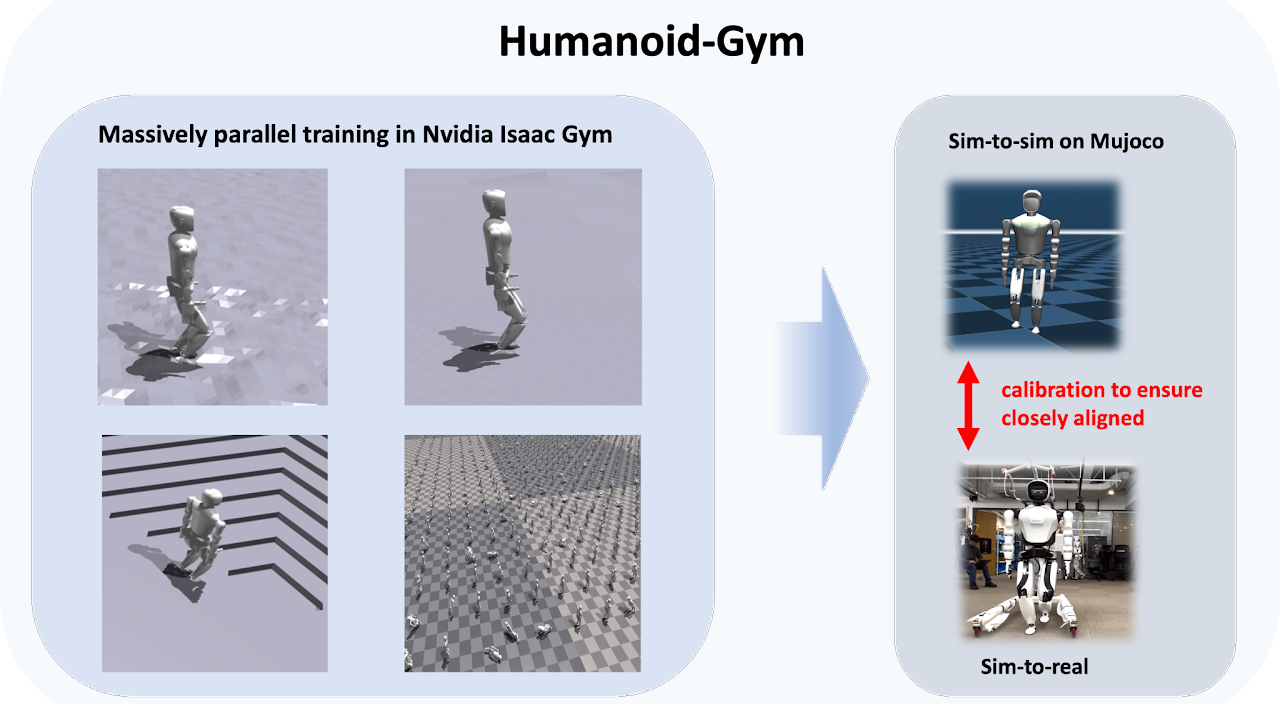}
    \caption{Pipeline of Humanoid-Gym. Initially, we employ massively parallel deep reinforcement learning (RL) within Nvidia Isaac Gym, incorporating diverse terrains and dynamics randomization. Subsequently, we undertake sim-to-sim transfer to test policies. Due to our meticulous calibration, the performance in both MuJoCo and real-world settings aligns closely.}
    \label{fig:pipeline}
\end{figure}

The workflow of Humanoid-Gym is illustrated in Fig.~\ref{fig:pipeline}. In this section, we will introduce the problem setting, system design, and reward design of our Humanoid-Gym.


\subsection{Reinforcement Learning For Robot Control}

Our approach employs a reinforcement learning model \(\mathcal{M} = \langle \mathcal{S}, \mathcal{A}, T, \mathcal{O}, R, \gamma \rangle\), with \(\mathcal{S}\) and \(\mathcal{A}\) denoting state and action spaces, \(T(\mathbf{s}'|\mathbf{s},\mathbf{a})\) the transition dynamics, \(R(\mathbf{s},\mathbf{a})\) the reward function, \(\gamma \in [0, 1]\) the discount factor, and \(\mathcal{O}\) the observation space. The framework is designed for both simulated and real-world settings, transitioning from full observability in simulations $(\mathbf{s}\in\mathcal{S})$ to partial observability in the real world (\(\mathbf{o}\in\mathcal{O}\)). This necessitates operating within a Partially Observable Markov Decision Process (POMDP) \cite{spaan2012partially}, with the policy \(\pi(\mathbf{a}|\mathbf{o}_{\leq t})\) mapping observations to action distributions to maximize the expected return \(J = \mathbb{E}[R_t] = \mathbb{E}\left[\sum_{t}\gamma^t r_t\right] \).

We leverage Proximal Policy Optimization (PPO) \cite{schulman2017proximal} loss, supplemented by the Asymmetric Actor Critic \cite{pinto2017asymmetric} method and the integration of privileged information during training, shifting to partial observations during deployment. The policy loss is defined as:
\begin{equation}
\label{eq:policy}
\begin{aligned}
\mathcal{L}_{\pi} &= \min \left[
\frac{\pi(a_t \mid o_{\leq t})}{\pi_{b}(a_t \mid o_{\leq t})} A^{\pi_{b}}(o_{\leq t}, a_t), \right. \\
&\quad \left. \text{clip}\left(\frac{\pi(a_t \mid o_{\leq t})}{\pi_{b}(a_t \mid o_{\leq t})}, c_1, c_2\right) A^{\pi_{b}}(o_{\leq t}, a_t)
\right]
\end{aligned}
\end{equation}

Advantage estimation utilizes Generalized Advantage Estimation (GAE) \cite{schulman2015high}, requiring an updated value function:

\begin{equation}
\label{eq:value}
\mathcal{L}_v = \| R_t - V(s_t)\|_2,
\end{equation}

\subsection{System design}




The base poses of the robot, denoted as \(P^b\), are six-dimensional vectors \([x, y, z, \alpha, \beta, \gamma]\), representing both the position coordinates \(x, y, z\) and the orientation angles \(\alpha, \beta, \gamma\) in Euler notation. The joint position for each motor is represented by \(\theta\), and the corresponding joint velocity by \(\dot{\theta}\). Furthermore, we define a gait phase \cite{siekmann2021sim,yang2022fast}, which comprises two double support phases (DS) and two single support phases (SS) within each gait cycle. The cycle time, denoted as \(C_T\), is the duration of one full gait cycle. A sinusoidal wave is employed to generate reference motion, reflecting the repetitive nature of the gait cycle involving pitch, knee, and ankle movements. Notably, we also designed a periodic stance mask \(I_p(t)\) (Fig \ref{fig:stance_mask}) that indicates foot contact patterns in synchronization with the reference motion. For instance, if the reference motion lifts the left foot, the right foot should be in the single support phase, with the foot contact mask indicated as $[0, 1]$; during DS phases, it would be $[1, 1]$.

The chosen action is the target joint position for the Proportional-Derivative (PD) controller. The policy network integrates proprioceptive sensor data, a periodic clock signal \([\sin(2 \pi t / C_T), \cos(2 \pi t / C_T)]\), and velocity commands \(\dot{P}_{x, y, \gamma}\). A single frame of input are elaborated in Table \ref{tab:observation}. Additionally, the state frame includes feet contact detect \(I_d(t)\) and other privileged observations.

\begin{table}[htp]
\centering
\caption{Summary of Observation Space. The table categorizes the components of the observation space into observation and state. The table also details their dimensions.}
\label{tab:observation}
\begin{tabular}{lccc}
\toprule
\textbf{Components} & \textbf{Dims} & \textbf{Observation} & \textbf{State} \\
\hline
Clock Input \((\sin(t), \cos(t))\) & 2 & \checkmark & \checkmark \\
Commands \((\dot{P}_{x, y, \gamma})\) & 3 & \checkmark & \checkmark \\
Joint Position $(\theta)$ & 12 & \checkmark & \checkmark \\
Joint Velocity $(\dot{\theta})$ & 12 & \checkmark & \checkmark \\
Angular Velocity $(\dot{P}^b_{\alpha \beta \gamma})$  & 3 & \checkmark & \checkmark \\
Euler Angle $(P^b_{\alpha \beta \gamma})$  & 3 & \checkmark & \checkmark \\
Last Actions $(a_{t - 1})$ & 12 & \checkmark & \checkmark \\
\hline
Frictions & 1 & & \checkmark \\
Body Mass & 1 & & \checkmark \\
Base Linear Velocity & 3 & & \checkmark \\
Push Force & 2 & & \checkmark \\
Push Torques & 3 & & \checkmark \\
Tracking Difference & 12 & & \checkmark \\
Periodic Stance Mask & 2 & & \checkmark \\
Feet Contact detection& 2 & & \checkmark \\
\bottomrule
\end{tabular}
\end{table}


Our control policy operates at a high frequency of $100$Hz, providing enhanced granularity and precision beyond standard RL locomotion approaches. The internal PD controller runs at an even higher frequency of $1000$Hz. For training simulations, Isaac Gym is utilized \cite{makoviychuk2021isaac}, while MuJoCo, known for its accurate physical dynamics, is chosen for sim2sim validation. This approach combines the benefits of high-speed GPU-based parallel simulation, albeit with less accuracy, and the high accuracy but slower CPU-based simulation.


The detailed settings for both algorithms and the environment designed are shown in Appendix TABLE \ref{tab:hyperparameters}. We use multi-frames of observations and privilege observation, which is crucial for locomotion tasks on uneven terrain.

\subsection{Reward Design}

Our reward function directs the robot to adhere to velocity commands, sustain a stable gait, and achieve smooth contact. The reward function is structured into four key components: (1) velocity tracking, (2) gait reward, and (3) regularization terms.

The reward function is summarized in Appendix Table \ref{tab:reward}. It is important to note that the commands $\text{CMD}_{z, \gamma, \beta}$ (velocity mismatch term) are intentionally set to zero. This is because we do not control them; rather, we aim to maintain their values at zero to ensure stable and smooth walking. In addition, the reward (contact pattern) encourages feet to align with their contact masks, denoting swing, and stance phases, as illustrated in Appendix Fig.~\ref{fig:stance_mask}. Therefore, the total reward at any time step $t$ is computed as the weighted sum of individual reward components, expressed as $r_t = \sum_i r_i \cdot \mu_i $, where $\mu_i$ represents the weighting factor for each reward component $r_i$.


\section{Experiments}
In this section, we will illustrate the result of zero-shot transfer for both sim-to-sim and sim-to-real scenarios. Additionally, we also provide visualization of the calibration for MuJoCo to verify the effectiveness of sim-to-sim. For the validation, we utilized Robot Era's humanoid robots, XBot-S and XBot-L, measuring 1.2 meters and 1.65 meters in height, shown in Appendix Fig.~\ref{fig:robots}, respectively.

\subsection{Zero-shot Transfer}

We carefully design domain randomization terms, as detailed in Appendix TABLE \ref{tab:domain_randomization}, to minimize the sim2real gap, following the approach outlined in \cite{tobin2017domain}. Our agents are capable of transitioning to real-world environments via zero-shot sim-to-real transfer, which is illustrated in Fig. \ref{fig:envs}. The standard procedure involves training agents on GPUs, followed by policy analysis in MuJoCo. For a comprehensive evaluation, we developed two types of terrains: flat and uneven, as depicted in Appendix Fig. \ref{fig:terrains}. The flat terrain replicates the environment encountered during training in Isaac Gym, while the uneven terrain offers a substantially more challenging landscape, differing significantly from our initial training scenarios. Remarkably, our trained policies enable the robots to traverse both types of terrain successfully.

\subsection{Calibration for MuJoCo}
\begin{figure}[h]
    \centering
    \includegraphics[width=1\linewidth]{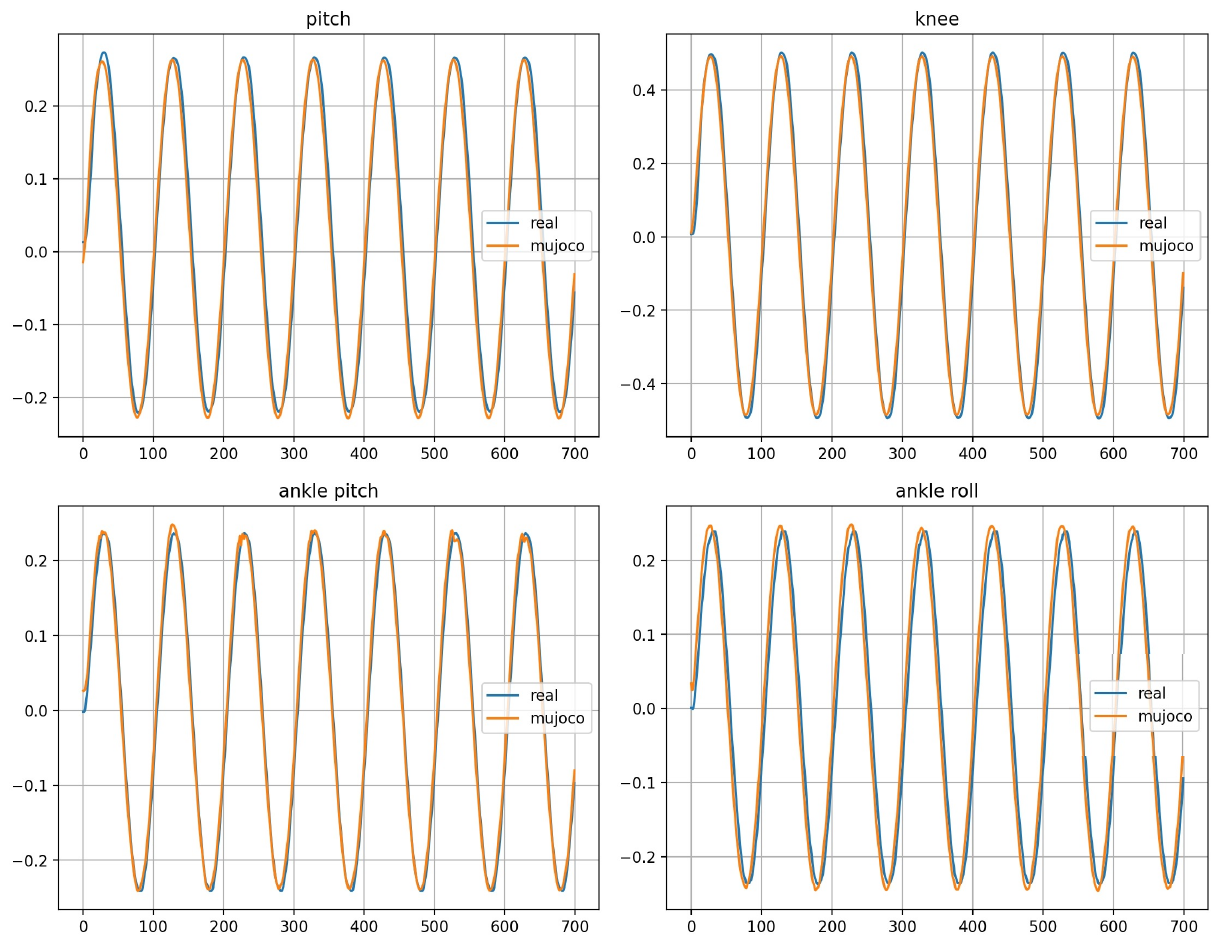}
    \caption{Sine wave in Both MuJoCo and real-world environment. It can be found that the trajectories of the two are very close after calibration.}
    \label{fig:sine}
\end{figure}

\begin{figure}[h]
    \centering
    \includegraphics[width=1\linewidth]{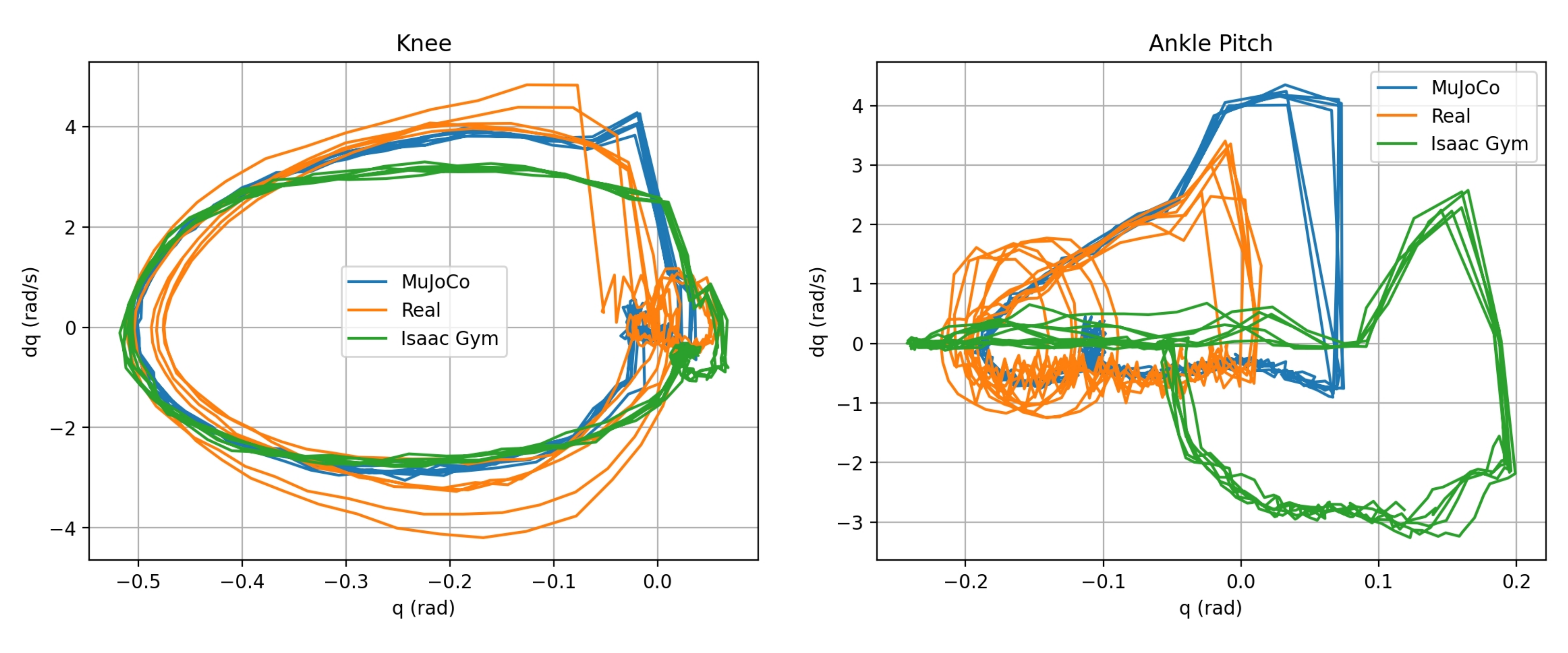}
    \caption{Phase Portrait for MuJoCo, Real-World Environment, and Isaac Gym.}
    \label{fig:portrait}
\end{figure}

We meticulously calibrated the MuJoCo environment to align its dynamics and performance more closely with that of the real world. By comparing the leg swing sine waves in both the MuJoCo and the real-world environments, which represent joint positions over time that track a specified sine wave, we observed nearly identical trajectories. These are depicted in Fig.~\ref{fig:sine}. Furthermore, we also compare the resulting phase portrait of the left knee joint and left ankle pitch joint\cite{radosavovic2024humanoid} within 5-second trajectories, 0.5 n/s heading velocity, as shown in Fig.~\ref{fig:portrait}. It is clear to see that the dynamics in MuJoCo are closer to the real environment than Isaac Gym.

\section{CONCLUSIONS}
Humanoid-Gym facilitates zero-shot transfer for humanoid robots of two distinct sizes, from sim-to-sim and sim-to-real, via a specialized reward function tailored for humanoid robotics. Our experimental outcomes reveal that the adjusted MuJoCo simulation closely mirrors the dynamics and performance of the real-world environment. This congruence enables researchers lacking physical robots to validate training policies through sim-to-sim, significantly enhancing the potential for successful sim-to-real transfers.


\addtolength{\textheight}{-2cm}   





\section*{ACKNOWLEDGMENT}
The implementation of Humanoid-Gym relies on resources from legged\_gym and rsl\_rl projects\cite{rudin2022learning} created by the Robotic Systems Lab. We specifically utilize the `LeggedRobot` implementation from their research to enhance our codebase.


\newpage
\bibliographystyle{IEEEtran}
\bibliography{IEEEexample}

\newpage
\appendix

\begin{figure}[hb]
    \centering
    \includegraphics[width=0.7\linewidth]{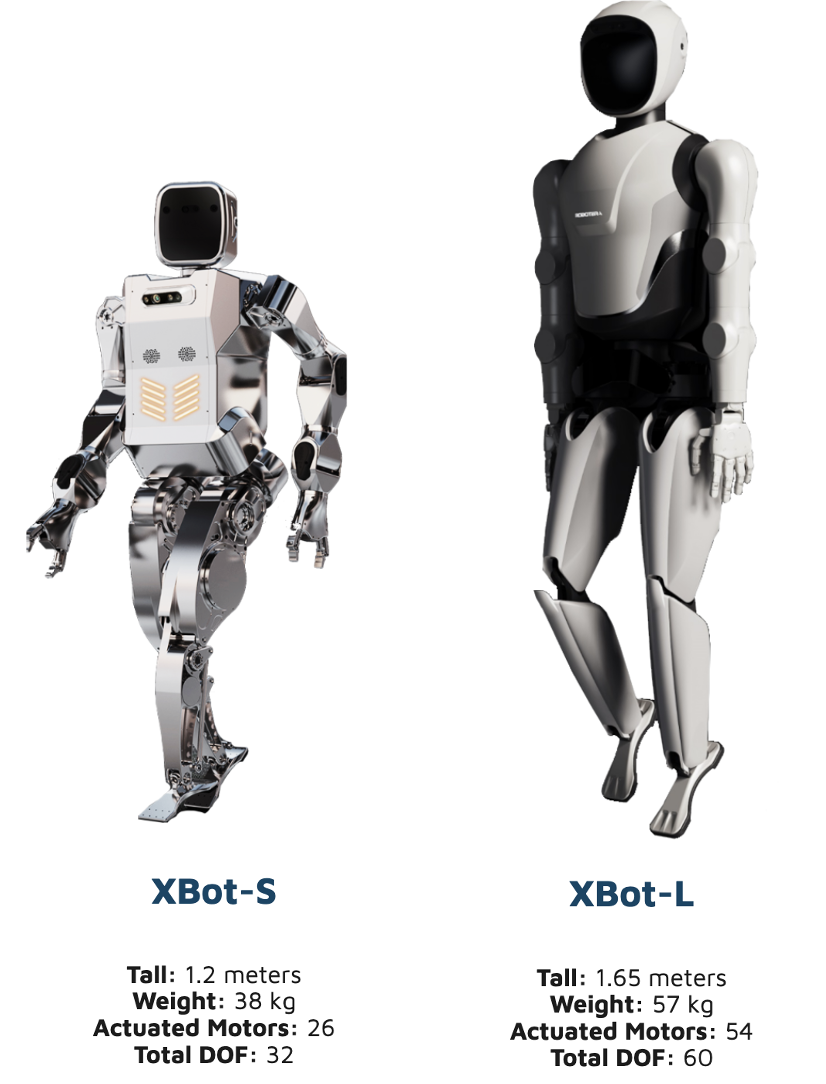}
    \caption{Hardware Platform. Our Humanoid-Gym framework is tested on two distinct sizes of humanoid robots, XBot-S and XBot-L, provided by Robot Era.}
    \label{fig:robots}
\end{figure}

\begin{figure}[hb]
    \centering
    \includegraphics[width=0.8\linewidth]{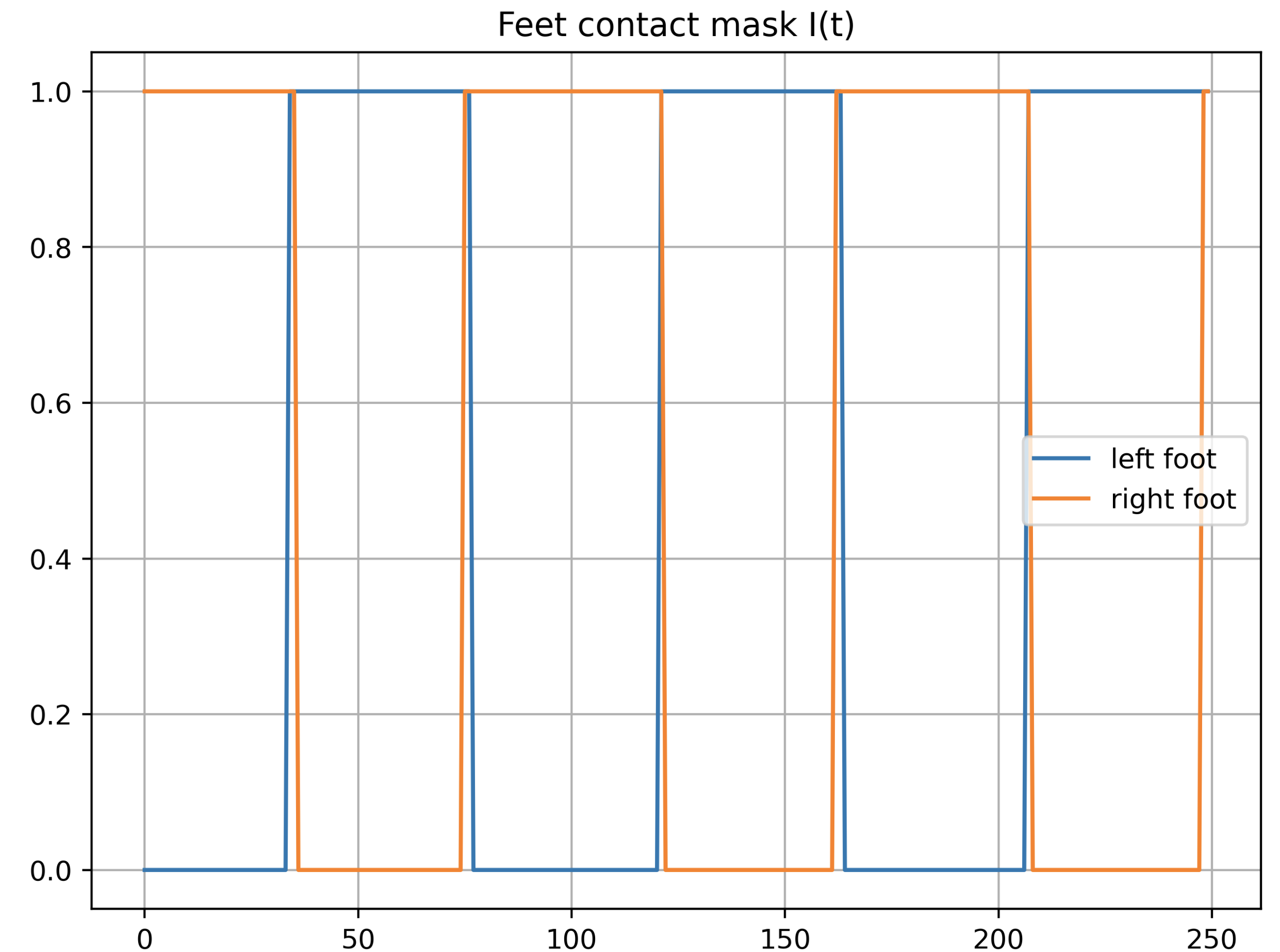}
    \caption{The stance mask is the contact planning for the left (L) and right (R) feet, where 0 indicates the swing phase and 1 indicates the stance phase is expected.}
    \label{fig:stance_mask}
\end{figure}

\begin{figure}[hb]
    \centering
    \includegraphics[width=0.9\linewidth]{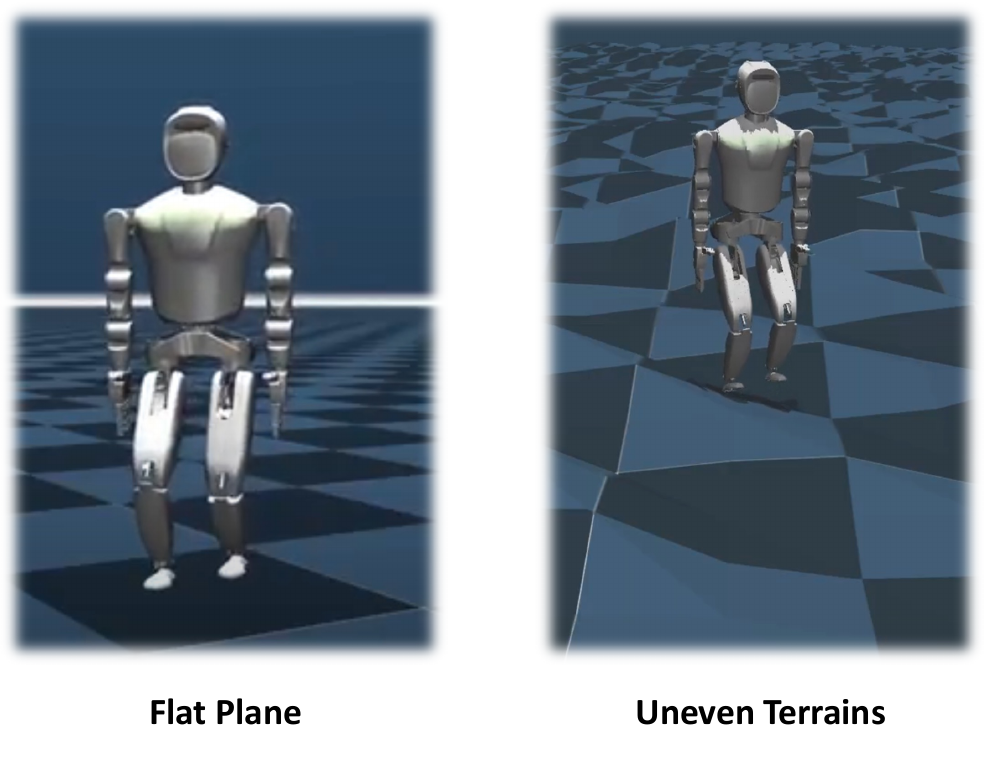}
    \caption{Terrains in MuJoCo. Humanoid-Gym provides two types of terrains utilized for sim-to-sim validation: flat planes and uneven terrains.}
    \label{fig:terrains}
\end{figure}

\begin{table}[hb]
\centering
\caption{Hyperparameters.}
\label{tab:hyperparameters}
\begin{tabular}{lc}
\toprule
\textbf{Parameter} & \textbf{Value} \\
\hline
Number of Environments & 8192 \\
Number Training Epochs & 2 \\
Batch size & $8192 \times 24$ \\
Episode Length & 2400 steps \\
Discount Factor & 0.994 \\
GAE discount factor & 0.95 \\
Entropy Regularization Coefficient & 0.001 \\
$c1$ & 0.8 \\
$c2$ & 1.2 \\
Learning rate & 1e-5 \\
\hline
Frame Stack of Single Observation & 15 \\
Frame Stack of Single Privileged Observation & 3 \\
Number of Single Observation & 47 \\
Number of Single Privileged Observation & 73 \\
\bottomrule

\end{tabular}
\end{table}

\begin{table}[hb]
\centering

\caption{Overview of Domain Randomization. Presented are the domain randomization terms and the associated parameter ranges. Additive randomization increments the parameter by a value within the specified range while scaling randomization adjusts it by a multiplicative factor from the same range.}

\label{tab:domain_randomization}
\begin{tabular}{lcccc}
\toprule
\textbf{Parameter} & \textbf{Unit} & \textbf{Range} & \textbf{Operator} & \textbf{Type} \\
\hline
Joint Position & rad & [-0.05, 0.05] & additive & Gaussian ($1 \sigma$) \\
Joint Velocity & rad/s & [-0.5, 0.5] & additive & Gaussian ($1 \sigma$) \\
Angular Velocity & rad/s & [-0.1, 0.1] & additive & Gaussian ($1 \sigma$) \\
Euler Angle & rad & [-0.03, 0.03] & additive & Gaussian ($1 \sigma$) \\
System Delay & ms & [0, 10] & - & Uniform \\
Friction & - & [0.1, 2.0] & - & Uniform \\
Motor Strength & \% & [95, 105] & scaling & Gaussian ($1 \sigma$) \\
Payload & kg & [-5, 5] & additive & Gaussian ($1 \sigma$) \\
\bottomrule
\end{tabular}
\end{table}

\begin{table}[hb]
\centering
\caption{In defining the reward function, we use a tracking error metric denoted by \(\phi(e, w)\). This metric is expressed as
\(
\phi(e, w) := \exp(-w \cdot \|e\|^2),
\)
where \(e\) represents the tracking error, and \(w\) is the associated weight. The target base height is set to \(0.7\,\text{m}\).
}
\label{tab:reward}

\begin{tabular}{@{}llr@{}}
\toprule
\textbf{Reward} & \textbf{Equation (\(r_i\))} & \textbf{reward scale(\(\mu_i\))} \\ \midrule

Lin. velocity tracking & $\phi(\dot{P}^b_{xyz} - \text{CMD}_{x y z}, 5)$ & 1.2 \\
Ang. velocity tracking &  $\phi(\dot{P}^b_{\alpha \beta \gamma} - \text{CMD}_{\alpha \beta \gamma}, 5)$ & 1.0 \\
Orientation tracking& \(\phi(P^b_{\alpha\beta}, 5)\) & 1.0 \\
Base height tracking & $\phi(P^b_{z} - 0.7, 100)$ & 0.5 \\
Velocity mismatch & $\phi(\dot{P}^b_{z, \gamma, \beta} - \text{CMD}_{z, \gamma, \beta}, 5)$ & 0.5 \\
\hline
Contact Pattern & \(\phi(I_p(t) - I_d(t), \infty)\)  &  1.0 \\
Joint Position Tracking & $\phi(\theta - \theta_\text{target}, 2)$ & 1.5\\
\hline
Default Joint & $ \phi(\theta_t - \theta_0, 2) $ & 0.2 \\ 
Energy Cost & \(|\tau||\dot{\theta}|\) & -0.0001 \\
Action Smoothness & $ \| a_t - 2a_{t-1} + a_{t-2}\|_2$ & -0.01 \\
Large contact & $ \text{max}(F_{L,R} - 400, 0, 100)$ & -0.01 \\
\bottomrule
\end{tabular}
\end{table}

\end{document}